\documentclass{article}

\usepackage{todonotes}
\usepackage{natbib}
\usepackage[utf8]{inputenc} 
\usepackage[T1]{fontenc}    
\usepackage{url}            
\usepackage{booktabs}       
\usepackage{amsfonts}       
\usepackage{nicefrac}       
\usepackage{microtype}      

\usepackage{amsmath}
\usepackage{commath}
\usepackage{graphicx}
\usepackage{placeins}
\usepackage{subfig}
\usepackage{float}

\usepackage[accepted]{icml2020}

\icmltitlerunning{A Graph VAE and Graph Transformer Approach to Generating Molecular Graphs}

\begin{document}

\twocolumn[
\icmltitle{A Graph VAE and Graph Transformer Approach to Generating Molecular Graphs}



\icmlsetsymbol{equal}{*}

\begin{icmlauthorlist}
\icmlauthor{Joshua Mitton}{cs}
\icmlauthor{Hans M. Senn}{chem}
\icmlauthor{Klaas Wynne}{chem}
\icmlauthor{Roderick Murray-Smith}{cs}
\end{icmlauthorlist}

\icmlaffiliation{cs}{School of Computing Science, University of Glasgow, United Kingdom}
\icmlaffiliation{chem}{School of Chemistry, University of Glasgow, United Kingdom}

\icmlcorrespondingauthor{Joshua Mitton}{j.mitton.1@research.gla.ac.uk}

\icmlkeywords{Machine Learning, ICML, Graph Neural Network, Graph Convolution, Transfromer, VAE, Generative Model, Chemistry, Molecules}

\vskip 0.3in
]

\printAffiliationsAndNotice{}  

\begin{abstract}
We propose a combination of a variational autoencoder and a transformer based model which fully utilises graph convolutional and graph pooling layers to operate directly on graphs. The transformer model implements a novel node encoding layer, replacing the position encoding typically used in transformers, to create a transformer with no position information that operates on graphs, encoding adjacent node properties into the edge generation process. The proposed model builds on graph generative work operating on graphs with edge features, creating a model that offers improved scalability with the number of nodes in a graph. In addition, our model is capable of learning a disentangled, interpretable latent space that represents graph properties through a mapping between latent variables and graph properties. In experiments we chose a benchmark task of molecular generation, given the importance of both generated node and edge features. Using the QM9 dataset we demonstrate that our model performs strongly across the task of generating valid, unique and novel molecules. Finally, we demonstrate that the model is interpretable by generating molecules controlled by molecular properties, and we then analyse and visualise the learned latent representation. 

\end{abstract}

\section{Introduction}
\begin{figure*}[t]
\vskip 0.2in
\begin{center}
\centerline{\includegraphics[width=1.2\columnwidth]{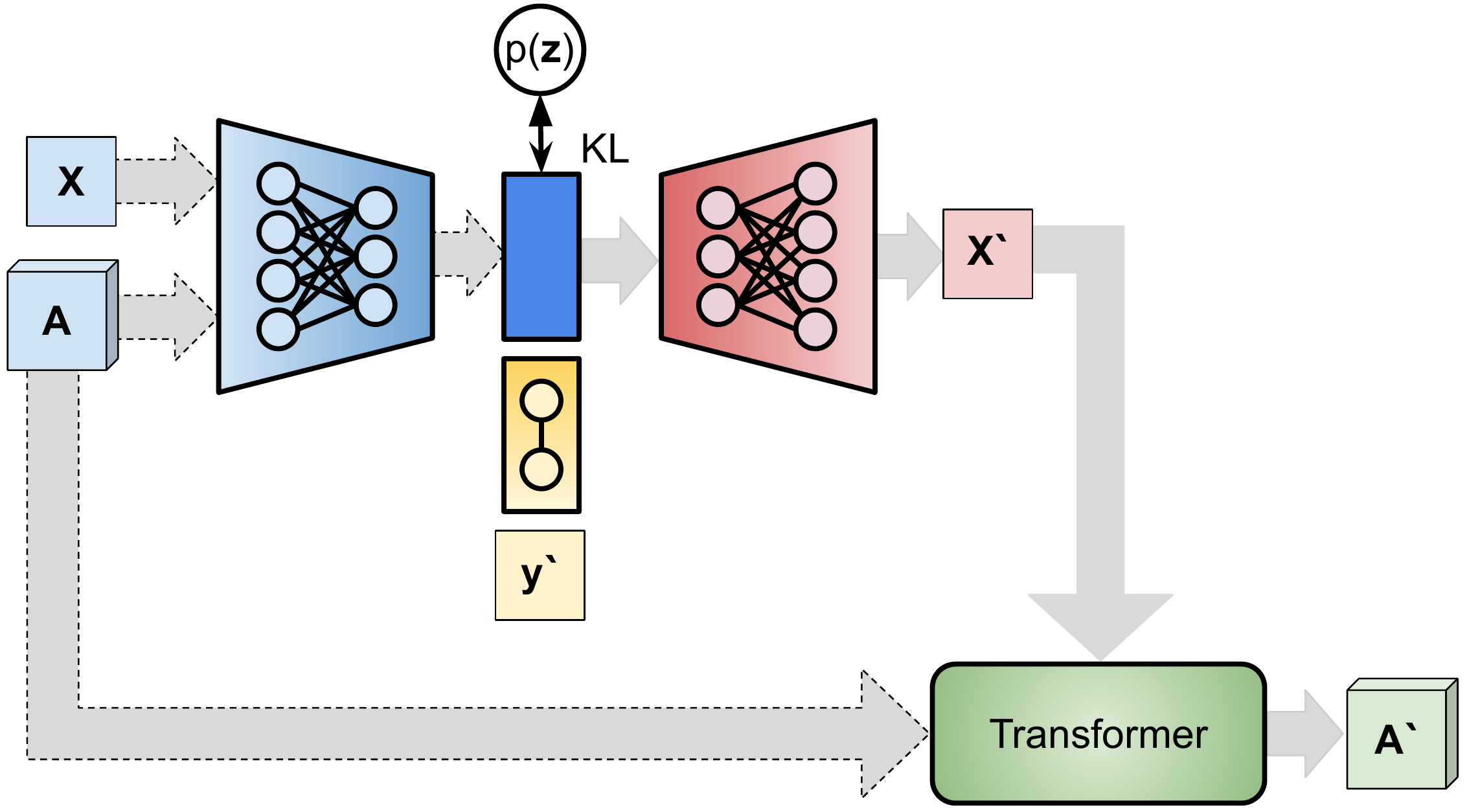}}
\caption{Illustration of the proposed VAE transformer model. The VAE takes in node feature matrix \textbf{X} and adjacency feature tensor \textbf{A}, and encodes these into a continuous latent representation \textbf{z}. Given a point in the latent space the VAE decoder outputs a node features matrix \textbf{X}'. In addition, graph level properties are predicted from a disentangled latent space, giving the graph property prediction \textbf{y}'. A transformer is used to make edge predictions, which takes in the adjacency feature tensor and predicted node feature matrix, and outputs a predicted edge feature tensor. When generating from the latent space arrows with dashed borders are removed}
\label{vae-transformer}
\end{center}
\vskip -0.2in
\end{figure*}

Designing chemical molecules with desired properties is a complex task, with the chemical space of small, carbon based molecules estimated to be in excess of $10^{60}$ structures \citep{dobson2004chemical}. Searching this space for molecules with desired properties is challenging given its discrete nature. Two popular methods to combat the discrete nature are: firstly, to convert molecules to SMILES \citep{weininger1988smiles}, a string-based representation, and secondly, to use deep learning methods that operate directly on graphs \citep{bronstein2017geometric, hamilton2017representation}. These  methods alleviate the challenge of the discrete nature of the molecular space. 

Using a SMILES representation of molecules has an issue with the one-to-many property between molecules and SMILES strings, which leads to a representation space in which molecules with similar properties are not embedded close together. Much of the recent progress in addressing the issue of molecular generation has made use of a SMILES based representation of molecules \citep{gomez2018automatic, kusner2017grammar, grisoni2020bidirectional}. On the other hand, training a model directly on a graph representation has demonstrated performance advantages when compared to converting graphs to a different less native representation \citep{kipf2016semi, velickovic2018graph}. Recent work for the generation of molecules has utilised the original graph representation of molecules \citep{de2018molgan, simonovsky2018graphvae}, but made use of a multi-layer perceptron (MLP) architecture for the graph generation, limiting the size of molecules that can be generated and the scalability of the model.

In this work we side-step these issues through the use of a graph convolutional decoder for generation of nodes and a transformer architecture utilizing a novel node encoding for edge prediction. This method builds on the variational graph auto-encoder \citep{kingma2013auto, kipf2016variational}; here a graph convolutional encoder and decoder are used along with differentiable graph pooling \citep{ying2018hierarchical} in a variational auto-encoder (VAE) to create graph level latent variables. In addition, a transformer architecture is used to predict edge features of the molecular graph, improving scalability over MLP-based approaches. Further, the transformer model has no concept of order, which is of benefit when learning about molecules. In addition, we introduce a node encoding that provides adjacent node information during edge prediction. Finally, in the VAE model we create a mapping from latent dimensions to graph properties. This enforces that molecules with specific properties are embedded close together in the latent space.

\section{Model}


\begin{figure}
\vskip 0.2in
\begin{center}
\centerline{\includegraphics[width=0.5\columnwidth]{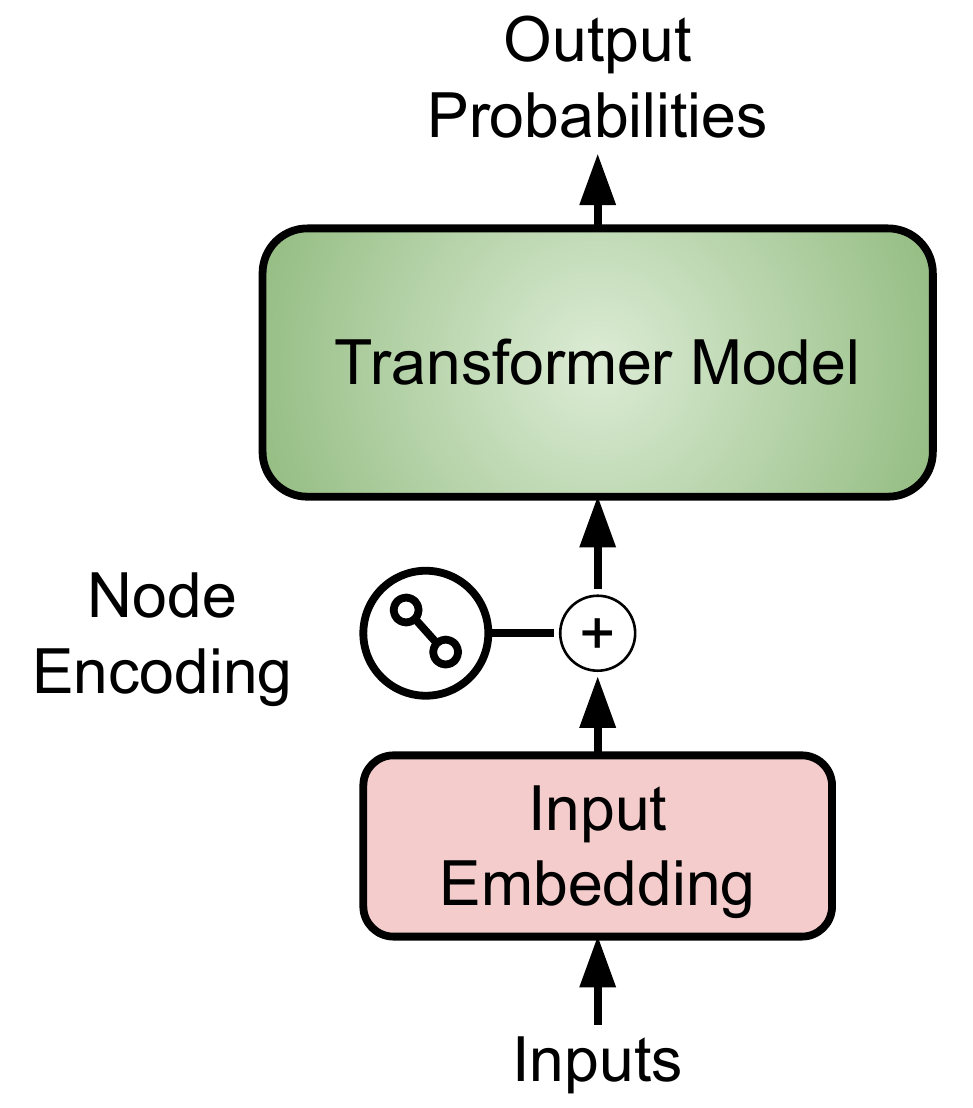}}
\caption{Illustration of the proposed transformer model. The model follows the typical layout of a transformer model, with the difference being the introduction of a novel node encoding. The node encoding injects information of adjacent nodes to the edge being predicted.}
\label{transformer}
\end{center}
\vskip -0.2in
\end{figure}

The model architecture is illustrated in Figure \ref{vae-transformer}, consisting of two main model components a VAE coupled with a transformer. The model takes as input a graph $\mathcal{G}$ representing a molecule and initially passes this to the VAE model. The model then outputs a graph in the form of node features from the VAE and edge features from the transformer.

\begin{table*}[t]
  \caption{Comparison of different models, without the use of expensive graph matching algorithms, on the QM9 dataset. Values are reported in percentages. Baseline results are taken from \citep{simonovsky2018graphvae}. Larger values are better.}
  \label{algorithm_qm9_1}
  \vskip 0.15in
  \begin{center}
  \begin{small}
  \begin{sc}
  \begin{tabular}{lcccc}
    \toprule
    Model               & Valid & Unique & Novel & Valid, Unique, and Novel \\
    \midrule
    CharacterVAE        & 10.3  & \textbf{67.5}   & 90.0  & 6.3  \\
    GrammarVAE          & 60.2  & 9.3    & 80.9  & 4.5  \\
    GraphVAE NoGM       & 81.0  & 24.1   & 61.0  & 11.9  \\
    MolGAN              & \textbf{98.1}  & 10.4   & \textbf{94.2}  & 9.6  \\
    \midrule
    GraphTransformerVAE & 74.6  & 22.5   & 93.9  & \textbf{15.8}  \\
    \bottomrule
  \end{tabular}
  \end{sc}
  \end{small}
  \end{center}
  \vskip -0.1in
\end{table*}

\subsection{VAE model}
The VAE takes as input a graph $\mathcal{G}$ representing a molecule and encodes this into a graph representation latent space $z \in \mathbb{R}^{c}$. The encoder is based on the GraphSAGE model \citep{hamilton2017inductive}, a framework shown to generate efficient node embeddings, and combines this framework with DIFFPOOL \citep{ying2018hierarchical}, a differentiable graph pooling module, and an MLP architecture to produce graph level latent representations of the input graph. The graph decoder reconstructs nodes from $c$-dimensional vectors $z \in \mathbb{R}^{c}$, sampling from a standard normal distribution $z \sim \mathcal{N} \left( \mathbf{0} ,\mathbf{I}\right)$. The graph decoder uses an MLP architecture, DIFFPOOL, and GraphSAGE based model in the opposite procedure to the graph encoder to unpool nodes from a single graph representation and predict reconstructed node features. Our model differs from MLP based approaches to graph generation in that it is only constrained to generating dense batches of graphs by the differentiable graph pooling/unpooling and, if replaced by a sparse version, the entire VAE model could operate on a sparse graph representation, which would further improve model scalability. In addition to the encoding and decoding modules, an additional molecular feature decoder is attached to the latent space that learns a mapping $h \left( z \right) \rightarrow{} \mu$ from latent variables $z$ to molecular properties similar to \citep{stoehr2019disentangling}. For the implementation of our model we make use of the PyTorch Geometric library \citep{Fey/Lenssen/2019}.

\subsection{Transformer model}
The transformer is used to make edge prediction in the graph, utilizing a sequential generation procedure. Here our model deviates from the traditional transformer model in that we do not make use of a position encoding to inject information about relative or absolute position into the model. Without the position encoding, the transformer model has no concept of order amongst the edges of the graph. There is no strict ordering in graphs, so we do not seek to introduce this order. Instead we introduce a novel node encoding into the embedding stage of the model which provides information about adjacent nodes, creating a transformer appropriate for edge prediction on graphs. Further information on the node encoding is provided in Appendix~\ref{nodeencode}. The implementation of our model is based on a vanilla transformer architecture \citep{opennmt}. At generation time the transformer sequentially adds edges to the graph, with the input being the node encoding of adjacent nodes to the edge being predicted and previously predicted edges in the graph. Further, the valency of an atom indicates the number of bonds that an atom can make in a stable molecule, and this imposes a constraint on generating valid molecules. Here we use a valency mask during the edge generation process to ensure the model does not predict a higher number of bonds than the valency of atoms, further valency mask information is provided in Appendix~\ref{valmask}. 
\begin{figure*}[t]
\vskip 0.2in
\begin{center}
\centerline{\includegraphics[width=1.5\columnwidth]{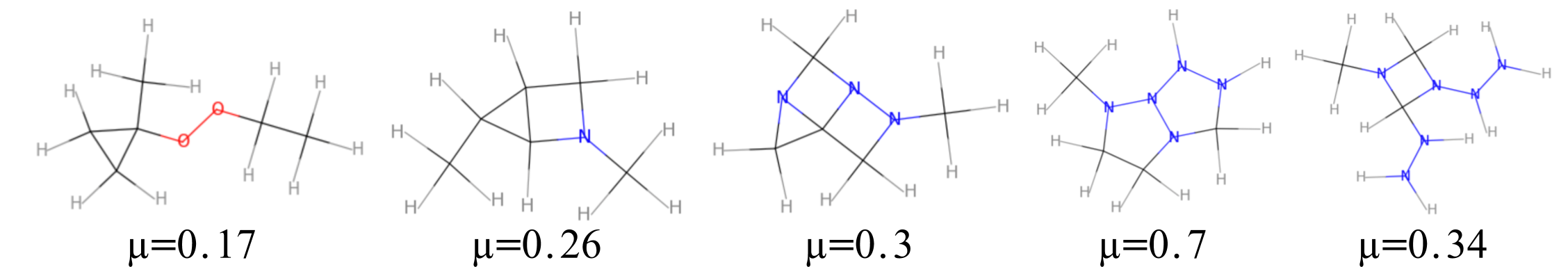}}
\caption{Generated molecules along a trajectory in the latent dimension disentangled for dipole moment, $\mu$~[D]. The trajectory is from a latent position of $-2$ through to $+2$ in steps of $1$.}
\label{fig:optimized_properties}
\end{center}
\vskip -0.2in
\end{figure*}

\section{Experiments: Molecule Generation}
We demonstrate how well our model performs on the task of molecule generation through the use of the QM9 dataset \citep{ramakrishnan2014quantum, ruddigkeit2012enumeration} and compare to previous results.

\subsection{Dataset}
The QM9 dataset \citep{ramakrishnan2014quantum, ruddigkeit2012enumeration} has become a benchmark task for molecule generation in past years. In all experiments we make use of this dataset, which is comprised of 133,885 organic compounds consisting of up to nine heavy atoms from carbon (C), nitrogen (N), oxygen (O), and fluorine (F). In addition, each molecule comes with a set of properties obtained from quantum-chemical calculations. Here we make use of the same decoder quality metrics as in \citep{simonovsky2018graphvae}, namely validity, uniqueness and novelty, which are explained in detail in the appendix.

In addition to decoding metrics, we analyse the graph latent space to evaluate the structure in relation to optimized graph properties. The ability to disentangle the latent space for molecular properties allows for control over the generation process, not only generating valid, unique or novel molecules, but molecules with specific quantum properties. 

\subsection{Results}

\subsubsection{Generating molecules}

Results are presented in Table~\ref{algorithm_qm9_1}, where the Graph Transformer VAE model is compared to baselines taken from the literature. It can be seen that our model scores strongly across all metrics validity, uniqueness and novelty, and this is more valuable than achieving state of the art in one score. 

We comparing our model to other models that do not make use of expensive graph matching algorithms, shown in Table~\ref{algorithm_qm9_1}. Here we only consider models that use a graph level embedding space. The initial measure of generating valid molecules is clearly important, as without this all generated molecules do not physically exist. Given this, all models outperforms CharacterVAE due to the low number of valid molecules this model generates. Aside from this model, our model is comparable with state of the art in all three measures or validity, uniqueness, and novelty. In addition, where our model is beaten in a specific measure by MolGAN or GraphVAE, we outperform them considerably in additional measures. Furthermore, our model achieves state of the art in generating valid, unique, and novel molecules demonstrating a step forward in overall molecular generation. In addition to performance on decoder metrics, the aim here is to develop a scalable model to larger graphs than those provided by the QM9 dataset. Here we reduce the growth of memory requirement with graph size from $\mathcal{O}(k^2)$ with GraphVAE to $\mathcal{O}(k)$ with our method, which could be further improved with the introduction of a differentiable pooling/unpooling operator on sparse graphs. 

\begin{figure}[ht]
\vskip 0.2in
\begin{center}
\begin{minipage}{0.65\columnwidth}
\centering
\subfloat[]{\label{main:a}{\includegraphics[width=\columnwidth]{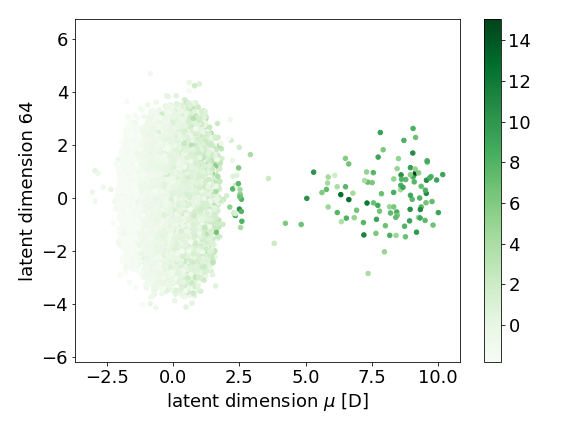}}}
\end{minipage}
\begin{minipage}{0.65\columnwidth}
\centering
\subfloat[]{\label{main:b}{\includegraphics[width=\columnwidth]{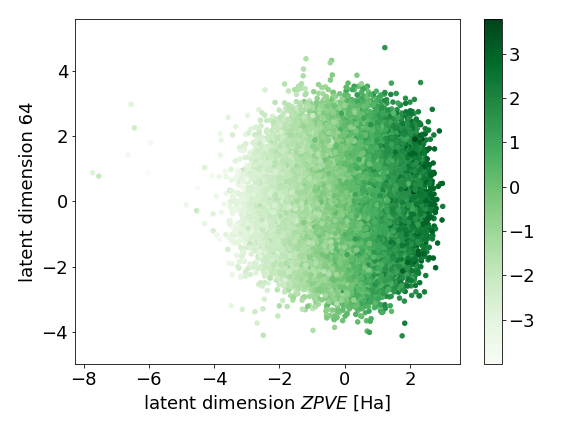}}}
\end{minipage}
\caption{Visualizations of disentangled latent space dimensions with disentangled axis on the $x$-axis and a dimension with no disentangled properties on the $y$-axis. Each figure is coloured by the molecular property that is disentangled. In (a) we demonstrate the disentangled property of dipole moment, $\mu$, and in (b) we demonstrate the disentangled property of zero point vibrational energy, $\mathrm{ZPVE}$.}
\label{fig:latentspace}
\end{center}
\vskip -0.2in
\end{figure}

\subsubsection{Directed molecule generation}

In addition to decoder quality metrics, we also demonstrate the quality of the disentangled latent space for creating molecules with desired properties. Here we show that the model has learned structure in the latent space for desired molecular properties, this makes the model useful for creating novel molecules with specific properties. This order in the latent space, allowing for controlled generation of new molecules, is essential if the model is to be used to find new useful molecules. Here we demonstrate in Figure~\ref{fig:latentspace} examples of the disentangled dimensions. 

Further, we demonstrate the capability of generating molecules with optimized properties in Figure~\ref{fig:optimized_properties} by generating five molecules uniformly spaced across the disentangled latent space dimensions. Here we chose to use the dipole moment, $\mu$, as this is the only quantum molecular property in the dataset that can be calculated with an approximate method. We use the package RDKit \citep{landrum2019rdkit} for calculations. Using Gasteiger partial charges \citep{gasteiger1980iterative}, $q$, and computing 3D coordinates \citep{riniker2015better}, $R$, the dipole moment is then calculated as $\abs{\mu} = \abs{\sum_{i \in N} R_{i} \times q_{i}}$, where $N$ is the number of atoms. The molecules generated show a structure gradually increasing until the final molecule, demonstrating capability of controlling the generation process. For control over this generation process having more order in the latent dimensions is better. As the disentangled latent dimension for dipole moment displays the least order, it will be the worst for controlling the generation process. Therefore, control over other quantum properties will be stronger.

\section{Conclusion}
We propose a generative graph based model capable of generating novel graphs, with both node and edge features, with desired graph level properties. The model improves the scalability of generative graph models, allowing the possibility of scaling the approach to the application of larger molecules. The disentangling of graph properties in the latent space provides control over the generative process and improves the interpretability of the model, allowing visualization of how individual graphs are positioned with respect to others for varying properties. The model scores strongly across three measures for molecular generation, namely validity, uniqueness and novelty, demonstrating state of the art molecular generation for the combined metric. Future work is required in the development of graph pooling to enable differentiable graph pooling/unpooling on sparse representations of graphs and in the construction of data-sets of larger molecules, with properties determined by the chemistry community that would be of importance, for example, for drug discovery and novel material development. 




\newpage

\bibliography{workshop_2020.bib}
\bibliographystyle{icml2020}

\newpage
\appendix




\section{Representation of the graph}
We consider the input graph $\mathcal{G}$, representing a molecule, to be comprised of nodes $\mathcal{V}$ and edges $\mathcal{E}$. Each atom equates to a node in the graph, given by $v_{i} \in \mathcal{V}$, which is expressed as a $\mathit{T}$-dimensional one-hot vector $\boldsymbol{x}_{i}$, indicating the type of atom. In addition, each bond equates to a edge in the graph, given by $e_{ij} \in \mathcal{E}$, which can either be expressed as a one-hot vector $\boldsymbol{y}_{ij}$ or bond type integer $y_{ij} \in \left \{ 1, ..., Y \right \}$, for $Y$ molecular bond types. This molecular graph can be expressed in a dense notation, with $N$ nodes, as a node feature matrix $\boldsymbol{X} = \left [ \boldsymbol{x}_{i}, ..., \boldsymbol{x}_{N} \right ]^{T} \in \mathbb{R}^{N \times T}$ and adjacency tensor $\boldsymbol{A} \in \mathbb{R}^{N \times N \times Y}$, where $A_{ij} \in \mathbb{R}^{Y}$ is a one-hot vector indicating the bond type between atoms $i$ and $j$ or $\boldsymbol{A} \in \mathbb{R}^{N \times N}$, where $A_{ij} \in \mathbb{R}$ is an integer indicating the bond type between atoms $i$ and $j$.

\section{Model Details}
The VAE encoder takes as inputs node feature matrix $\boldsymbol{X} \in \mathbb{R}^{N \times T}$ and adjacency tensor $\boldsymbol{A} \in \mathbb{R}^{N \times N \times Y}$. The VAE predicts a dense representation of nodes, which, when batched, has shape $\textbf{X}' \in \mathbb{R}^{B \times N \times T}$, where $B$ is the batch size, which scales linearly with graph size. The transformer model comprises of an encoder, decoder, embedding, and generator. The transformer model outputs an adjacency tensor with probabilistic edge type.

\section{Model training}
Here we reserve 10,000 molecules for testing, 10,000 of molecules for validation, and use the remaining molecules for training. 

The model is trained in two stages: training of the VAE and then training of the VAE and transformer. The first stage sees the VAE trained to generate node features and disentangle graph properties in the latent space. During this first stage, the objective function is $\mathcal{L} = \lambda_{0} \mathcal{L}_{n} + \lambda_{1} \mathcal{L}_{KL} + \lambda_{2} \mathcal{L}_{h},$ where $\lambda_i$ are scaling constants, $\mathcal{L}_{n}$ is the cross entropy loss of reconstructing node features, $\mathcal{L}_{KL}$ is the KL divergence loss, and $\mathcal{L}_{h}$ is the mean squared error loss of predicting graph properties from the latent space. During the second stage, we begin training the transformer, and the loss function is the same as during the first stage, with the addition of $\lambda_{3} \mathcal{L}_{t}$, where $\lambda_{3}$ is a scaling constant and $\mathcal{L}_{t}$ is the cross entropy loss of reconstructing edge features. Training is done with a batch size of 32; the first stage is trained for 70 epochs, then the second stage is trained for a further 650 epochs. The VAE model uses an Adam optimizer with initial learning rate of $5 e^{-4}$ and weight decay of $1 e^{-5}$; we also use a scheduler scaling the step size by $\nicefrac{1}{5}$ every $20$ epochs. The transformer model also uses an Adam optimizer, but the learning rate is a function of the number of epochs completed since the initial epoch the transformer begins training. 

\section{Evaluation Metrics}
Here we make use of the same decoder quality metrics as in \citep{simonovsky2018graphvae}, namely validity, uniqueness and novelty. Let $V$ be the list of chemically valid molecules generated and $n$ be the number of molecules generated. Then validity is defined as the ratio between the number of chemically valid molecules generated and the number of molecules generated. We define validity as the ability to parse the graph to a SMILES string using the RDKit parser \citep{landrum2019rdkit}. Further, uniqueness is defined as the ratio between the number of unique samples in the list $V$ and the number of valid samples. Finally, novelty is defined as the fraction of novel out-of-dataset graphs; this is the fraction of unique molecules generated that do not exist in the original training dataset.

\section{Valency Mask} \label{valmask}
A valency mask is used to allow the model to prioritize the prediction of the correct bond between atoms. The valency mask takes the predicted adjacency matrix and sequentially removes bonds between atoms if the number of bonds exceeds the valency of either atom.

\section{Node Encoder} \label{nodeencode}
The node encoder is used in the transformer model to inject information about adjoining nodes when predicting an edge type. Without this the transformer would only have information about edges that have already been predicted in the graph. When predicting molecular graphs the nodes/atoms are of importance when predicting the edge/bond between them. Therefore, the node encoding injects information of adjacent nodes to the edge being predicted into the transformer model. Here we utilize a node encoding that consists of a two layer MLP as the node encoding, but the specifics of the node encoding architecture can be modified depending on the task.

\newpage

\section{Further Disentangled Latent Dimensions}
\begin{figure}[ht]
\vskip 0.2in
\begin{center}
\begin{minipage}{0.55\columnwidth}
\centering
\subfloat[]{\label{main:c}{\includegraphics[width=\columnwidth]{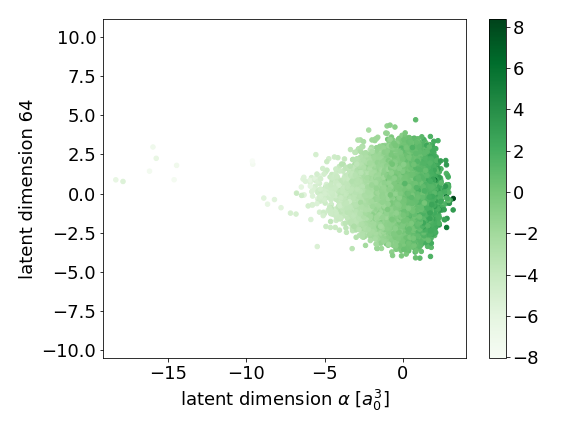}}}
\end{minipage}
\begin{minipage}{0.55\columnwidth}
\centering
\subfloat[]{\label{main:d}{\includegraphics[width=\columnwidth]{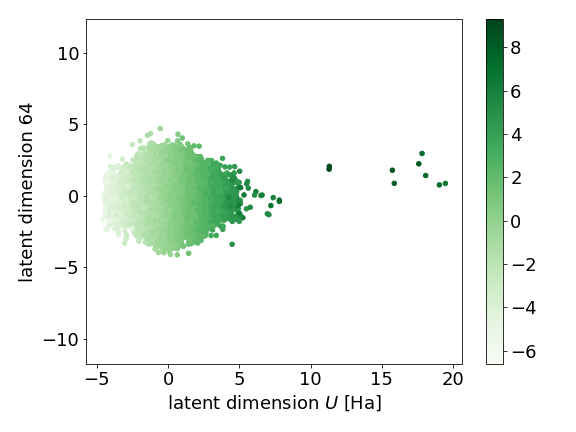}}}
\end{minipage}
\caption{Visualizations of a further two of sixteen disentangled latent space dimensions, with disentangled axis on the $x$-axis and a dimension with no disentangled properties on the $y$-axis. Each figure is coloured by the molecular property that is disentangled. In (a) we demonstrate the disentangled property of isotropic polarizability, $\alpha$, and in (b) we demonstrate the disentangled property of internal energy at 298.15~K , $U$.}
\label{fig:latentspace2}
\end{center}
\vskip -0.2in
\end{figure}


\end{document}